\documentclass[a4paper]{article}
\usepackage{spconf}

\usepackage{booktabs} 
\usepackage{color} 
\usepackage{multirow} 
\usepackage{graphicx}
\usepackage{amssymb,amsmath,bm,graphicx,url}
\usepackage{textcomp}
\usepackage{subcaption}
\usepackage{enumitem}
\usepackage[noadjust]{cite}
\usepackage{pgfplots}
\usepackage{filecontents}
\pgfplotsset{compat=1.13}

\ninept

\title{Lessons from Building Acoustic Models with a Million Hours of Speech}
\makeatletter
\def\name#1{\gdef\@name{#1\\}}
\makeatother
\name{{ \em Sree Hari Krishnan Parthasarathi,  Nikko Strom}}

\address{Amazon.com, USA.\\
{\small \tt \{sparta,nikko\}@amazon.com}}

\begin{document}
\maketitle
\begin{abstract}
This is a report of our lessons learned building acoustic models from 1 Million hours of unlabeled speech, while labeled speech is restricted to 7,000 hours. We employ student/teacher training on unlabeled data, helping scale out target generation in comparison to confidence model based methods, which require a decoder and a confidence model. To optimize storage and to parallelize target generation, we store high valued logits from the teacher model. Introducing the notion of scheduled learning, we interleave learning on unlabeled and labeled data. To scale distributed training across a large number of GPUs, we use BMUF with 64 GPUs, while performing sequence training only on labeled data with gradient threshold compression SGD using 16 GPUs. Our experiments show that extremely large amounts of data are indeed useful; with little hyper-parameter tuning, we obtain relative WER improvements in the 10 to 20$\%$ range, with higher gains in noisier conditions.
\end{abstract}


\section{Introduction}
\label{sec:intro}

A well-known maxim in the speech community is {\it there is no data like more data}~\cite{jelinek:2004}. Increasing the size of the training data by an order of magnitude have consistently led to substantial improvements in accuracy (\cite{moore2003comparison}, \cite{chelba2013}). In this paper we push the envelope in building acoustic models (AM) on extremely large amounts of data. Specifically, we report our lessons in building acoustic models on 1 Million hours of unlabeled speech, while using only 7,000 hours of labeled speech data.

Taking a historic perspective; in the 1920's Radio Rex\footnote{Arguably the first speech recognition system~\cite{jurafsky:2009}.} used thresholds on formant energies to recognize the word ``Rex''. Since then, automatic speech recognition (ASR) systems have become ever more complex and used increasing amounts of speech data.  Over the decades, corpora have grown from a few tens of hours of speech (TIDIGITS~\cite{tidigits:1993}, TIMIT~\cite{timit:1993}, WSJ~\cite{wsj:1992}), to a few hundred hours (Switchboard~\cite{switchboard:1992}), to a few thousand hours of speech (Fisher corpus~\cite{fisher:2004}).
In symbiosis with this growth of data and more powerful computing hardware, a similar evolution in model complexity and algorithms can be traced, from the hard-wired analog signal processing of Radio Rex, via template based pattern matching and dynamic time-warping \cite{sakoedynamic:1978}, to hidden Markov models \cite{rabinertutorial:1989}, and the current prevalent deep neural networks \cite{hintondeep:2012}.

Recently, with deep learning models, training data sizes on the order of ten thousand hours of speech are not unusual (\cite{deepspeech2:2016}, \cite{gong-semi:2016}). Building an AM from a hundred thousand hours is still rare, but~\cite{soltau:2016} showed that increasing from several thousand hours of training data to a hundred thousand hours of lightly supervised data can yield substantial accuracy improvements. As a note, the large dataset used in our work is fully unlabeled.

Semi-supervised learning (SSL) has a long history in ASR (\cite{kemp1999unsupervised}, \cite{lamel2002lightly}, \cite{ma2006unsupervised}).
Self-training is the most commonly used approach where typically there is a smaller labeled dataset, and a much larger unlabeled dataset.
The labeled data is used to train a seed model from a powerful model family, which is used to decode the unlabeled data at the second stage (often large beam sizes are used). The most reliable hypotheses are selected based on confidence measures~\cite{siu1997improved} and the speech data with the selected hypotheses are used for re-training the AM.

Self-training requires good confidence measures, which has been a challenge for SSL~(\cite{siu1997improved}, \cite{huang2013semi}). Several methods to estimate word and frame level token confidence from speech lattices or hypotheses have been developed (\cite{thomas2013deep}, \cite{liao2013large}). With models that have high memorization capability such as LSTM AMs, label quality becomes even more important~\cite{huang2016semi}. Another challenge for the scale of data we consider in this paper, is an efficient inference mechanism to not only generate lattices/hypotheses, but also to estimate token confidence and use it for hypothesis selection. A further challenge is applying sequence discriminative training, where label errors have a larger detrimental effect (\cite{manohar2015semi}, \cite{huang2008maximum}).

We built an SSL infrastructure that can train models on 1 Million hours of audio with a quick turnaround time. This paper reports our lessons in terms of the design choices made while building models at this scale. We based our training on the student/teacher paradigm. Recently, student/teacher training has become popular in the speech community for model compression ~(\cite{lu2017knowledge}, \cite{tucker2016model}, \cite{Li:Domain-adapt}). Here, instead student/teacher training is applied to produce soft targets for unlabeled data, which leads to efficient target generation ~(\cite{ba2014deep}, \cite{hinton2015distilling}). Further, we introduce a particular learning schedule -- interleaving training on labeled data with training on unlabeled data. Sequence training is also used, but only using labeled data. Finally, in our large-scale experiments we contrast two types of distributed training.

The remainder of the paper is organized as follows: starting with a description of the baseline fully supervised AM system in Section~\ref{sec:sys_description}, we discuss the semi-supervision design choices in Section~\ref{sec:ssldesign}. Next, we cover the experimental setup in Section~\ref{sec:exp_setup}, and validation results exploring the design choices in Section~\ref{sec:validation_results}. The final 1 Million hour results with analyses are described in Section~\ref{sec:final_results}, followed by our conclusions in Section~\ref{sec:conclusions}.
\section{Baseline Supervised Acoustic Model}
\label{sec:sys_description}
We use an HMM-LSTM hybrid. The HMM models low-frame rate single state triphone units~\cite{pundak2016lower}. States are clustered down to 3,183 senones using phonetic decision trees. The acoustic features consist of 64-dimensional log mel-warped energies computed on audio signals every 10 ms with a 25 ms analysis window~(\cite{parthasarathi2015fmllr},~\cite{garimella2015robust}). These are stacked three at a time and sub-sampled to a 30 ms advance. A causal mean estimate is computed and subtracted, and finally global mean and variance normalization is applied. To compensate for sub-sampling, features are created at three different offsets for each utterance.

The LSTM model is a stack of five layers, each consisting of 768 units resulting in about 24 M parameters. The model has a three-frame look-ahead. The training data is 7,000 hours of labeled US English data drawn from the Echo family of devices. The models are trained first with the cross-entropy criterion (CE), using alignments computed on the labeled data. First, we follow an exponential learning rate decay for ten epochs, with chunked BPTT for greater parallelization efficiency~\cite{doetsch2014fast}. In this technique, utterances are split into smaller sub-sequence chunks (here, 32 frames) and the sub-sequences are randomized. For each epoch we cycle through a different feature offset. Then the models are fine-tuned using full sequence CE BPTT for two more epochs. Finally, three epochs of the sequence discriminative criterion state-level minimum Bayes risk (sMBR) is applied.

We employ distributed training using synchronous SGD on two p3.16xlarge instances (16 Tesla V100 GPU cards).
Gradient Threshold Compression~\cite{strom2015scalable} is used for efficient peer-to-peer weight updates after every minibatch.

\section{Large-Scale Semi-Supervised Learning}
\label{sec:ssldesign}
At the scale of 1 Million hours, certain design choices were crucial for experiment turnaround time, while also obtaining significant accuracy improvements. This section presents various design choices and their considerations.

\subsection{Data Selection and Feature Extraction}
We drew data according to a device distribution roughly similar to that of the labeled data. Within each device, we drew samples randomly from the production data firehose. We did not filter data with confidence models nor for background speech/noise. Our hypothesis was that well-calibrated posteriors from the teacher model would mitigate poorly selected data.

To speed up parallel feature generation we did not require a pre-roll of utterances for initialization as described in~\cite{king2017robust}. We developed a feature pipeline that uses an efficient hashing mechanism to cluster speakers and sort utterances belonging to a speaker for performing running cepstral mean normalization. This could then be parallelized over several thousand CPU cores.

\subsection{Student-Teacher Learning}
A key design choice was to employ the student/teacher learning paradigm, thus taking the ASR decoder out of the SSL recipe. In essence, for each feature vector, the teacher network outputs a probability distribution over senones. The student network also estimates the probabilities over the senones given the same feature vector, and the learning objective optimizes the CE loss between these two distributions. The student models are identical to the LSTMs described in the previous section, but the teacher models have five bi-directional LSTM layers, each of size 768 units (amounting to a total of 78 M model parameters). The training of the teacher on the labeled data follows the same recipe as the regular LSTMs, discussed in Section~\ref{sec:sys_description}.

\subsubsection{Confidence Modeling}
There is evidence that even unfiltered data can lead to significant SSL improvements (\cite{lamel2002lightly}, \cite{lamel2002unsupervised}). Further, as neural networks have improved, the estimated probabilities become better calibrated~\cite{guo2017calibration}. Our hypothesis was that the teacher's posteriors are calibrated well enough to act as the confidence measure for the student training. However, in a traditional self-learning system, the language model is also providing additional information during the decoding, which is not present in our system. We hypothesize that this is partially mitigated by the bi-directional LSTM model, which has more context than the student.

\subsubsection{Target Generation}
The senone output distribution is large, and generating targets from the teacher model on-the-fly can slow down training. To reduce bandwidth and storage requirements as we parallelize across multiple GPUs, we store only the $k$ highest valued logits. During the student model training, full posteriors are reconstructed by filling the missing logits with large negative values. While this reconstruction is lossy, we found empirically that the probability mass is dominated by the top few posteriors. We found storing the top-20 values for $k$ to be sufficient from the standpoint of not having a WER degradation, while yielding a huge gain in storage. 

\subsection{Scheduled Learning}
While we primarily train on unlabeled data, the limited labeled data is also used. Learning on unlabeled and labeled data is interleaved, with slightly higher learning rates on the labeled data.

We used two unlabeled training datasets (100khrs and 1Mhrs), as will be discussed in Section~\ref{sec:exp_setup}. Given the large amounts of data, our design was to perform just one learning pass through the data. We divided the data into a number of \textit{sub-epochs}, with a sub-epoch defined as 25,000 and 55,000 hours for the 100khr and 1Mhr datasets respectively. We decayed the learning rate as we passed through the sub-epochs, following an exponential learning rate decay.

For the 100khrs, after each sub-epoch through the unlabeled data, we perform CE training on the labeled data, with a rotation through the feature offsets (refer to Section~\ref{sec:sys_description}). For the 1Mhr data, after every five sub-epochs through the unlabeled data, we perform CE training on the labeled data, rotating through the feature offsets.

As discussed in Section~\ref{sec:sys_description} we employ sequence chunked BPTT for training speed. On the 100khrs set, chunked training is used for the first three sub-epochs (including the corresponding passes through the labeled data), followed by a full sequence BPTT on the last sub-epoch on the unlabeled data. On the 1Mhrs data, we apply chunked training for the first 15 sub-epochs, and then do fine-tuning during the last three sub-epochs.

\subsection{Sequence Training for SSL}
Sequence discriminative training often yields large accuracy gains (commonly, around 10$\%$ relative). However, it is also a difficult problem for SSL~(\cite{huang2010semi}, \cite{manohar2015semi}), since it is particularly sensitive to noisy references during training. We chose to perform sequence training only on labeled data. There was evidence ~\cite{kanda2016investigation} that the accuracy gains may be relatively small in such a setup. However, our hypothesis was that this result was due to a smaller labeled dataset, and using our full 7,000 hour labeled data would still recover large gains from sequence training.

\subsection{Distributed Training}
For the scale of data we want to learn from, our design goal was to parallelize beyond a few tens of GPUs. We explored the Gradient Threshold Compression method (GTC)~\cite{strom2015scalable} and Blockwise Model-Update Filtering (BMUF)~\cite{bmuf-2016}.

With high-end GPUs like Tesla V100s, gradient compression based training scales well up to 16 GPU cards, but efficiency tapers off at higher scale. In this work, we used two p3.16xlarge instances (16 Tesla V100 GPU cards spread over two hosts).

The BMUF training scales nearly linearly with GPUs, at least in terms of throughput, because the per-worker model updates happen much more infrequently. However, it can come at a cost in accuracy. The Nesterov-like momentum updates at block level recover some of these losses~\cite{li2017empirical}, but we still see some degradation (Table \ref{tab:tab_sequence}). For our experiments with BMUF we used eight p3.16xlarge instances (64 Tesla V100 GPUs).

\section{Experimental Setup}
\label{sec:exp_setup}
We discussed a number of system level details in Section~\ref{sec:sys_description}. In this section we give the details with regard to our experimental setup.

\subsection{Training Datasets}
For our experiments we used three far-field training datasets drawn from production data of the Alexa family of devices from the US English locale: (a) a 7,000 hour fully labeled dataset (b) 100,000 hours of unlabeled data for prototyping and validating design choices, and (c) a 1 Million hour unlabeled dataset for the final model build.

\subsection{Test Datasets}
We used several test sets in this work: (a) a validation test set (referred to as VAL), which consisted of about 30 hours of data, (b) acoustically difficult audio data collected in a real room with about 5,000 utterances roughly equally spread among five device placements. The first device placement (DP1) in the center of the room was the easiest, while other conditions (DP2 to DP5) were more challenging, and (c) a 30 hours independent test set (referred to as TEST). The TEST set was also divided into native (TST-NATIVE) and non-native (TST-NON-NAT) speakers as judged by the annotators.

\subsection{Decoding Setup and Scoring}
All decoding on the VAL test set use a 4-gram statistical language model (LM). The acoustic model scale factor was tuned on this test set. For the decoding runs on all other test sets, the statistical LM was combined with a set of domain-specific grammars. We report results as relative Word Error Rate Reduction (WERR) compared the strong baseline supervised learning system.

\section{Experiments on 100,000 Hours}
\label{sec:validation_results}
\vspace{-1mm}
In this section, we validate the key design choices by training models on the 100,000 hour unlabeled data and decoding on the VAL test set.
The key elements are: (a) scheduled learning and its interaction with sMBR trained teacher, (b) sequence training of the student model, and (c) choice of distributed training method.


\subsection{Scheduled Learning}
We perform our analysis with and without scheduled learning; we also consider its interaction with and without sMBR trained teacher. Table~\ref{tab:tab_scheduled} presents accuracy for the four different options relative to a baseline LSTM AM that is trained with the CE criterion on the fully labeled 7,000 training data.

\begin{table}[t]
\centering
\caption{\textit{On VAL test set, relative WER ($\%$) reduction for SSL student models trained on 100,000 hour dataset: with and without scheduled learning (SL); with and without sMBR trained teachers. The WER reduction is computed against a baseline LSTM AM that is trained with CE criterion on the fully labeled 7,000 hour training data.}}
\renewcommand{\arraystretch}{1.3}
\begin{tabular}{|c|c|c|c|c|c|}
  \hline
 & without sMBR teacher & with sMBR teacher  \\
\hline
without SL &  $1.0$ & $8.8$ \\
\hline
with SL & $6.8$ & $\mathbf{10.8}$ \\
\hline
\end{tabular}
\vspace{-4mm}
\label{tab:tab_scheduled}
\end{table}

It can be seen from the table that scheduled learning, i.e., interleaving labeled data in the learning, helps the student models both in the case of CE trained as well as sequence trained teachers. However, the gain with scheduled learning is more with students trained with the CE-teacher.


\subsection{Scaling Number of GPUs to 64}
Student models used in Table~\ref{tab:tab_scheduled} were trained using the GTC trainer with 16 GPUs. Using the best model configuration from Table~\ref{tab:tab_scheduled}, i.e., with scheduled learning and with sMBR trained teachers, we now investigate the effect of BMUF trainer on student models. For student training with BMUF trainer, we use 64 GPUs. Note that the objective here is not to compare the BMUF and the GTC trainers (which would involve an extensive search over hyper-parameters of both trainers), but to obtain an estimate of the WER gain or loss in scaling up the number of GPUs in training (for which we use BMUF). With 64 GPUs, we obtain a relative WER reduction of 7.8\% over the baseline LSTM AM that is trained with the CE criterion on the fully labeled 7,000 training data (compared to 10.8\% in Table~\ref{tab:tab_scheduled}). Thus, in attempting to scale to 64 GPUs, we lose some of the gains due to SSL.

\begin{table}[ht]
\centering
\caption{\textit{On VAL test set, relative WER ($\%$) reduction for sequence training of SSL students. The WER reduction is computed against a baseline LSTM AM that is trained with CE criterion on the fully labeled 7,000 hour training data. sMBR is performed with GTC trainer.}}
\renewcommand{\arraystretch}{1.3}
\begin{tabular}{|c|c|c|c|c|c|}
  \hline
 System & CE Trainer &WERR ($\%$)  \\
\hline
Baseline labeled CE & GTC & $0$\\
\hline
Baseline labeled CE $+$ sMBR & GTC & $10.7$\\
\hline
SSL $+$ SL $+$ sMBR & GTC & $18.6$  \\
\hline
SSL $+$ SL $+$ sMBR & BMUF &  $15.6$  \\
\hline
\end{tabular}
\vspace{-4mm}
\label{tab:tab_sequence}
\end{table}

\subsection{Sequence Training}
Recall that our strategy is to perform sequence training of student models only on the 7,000 hour labeled dataset. We compare if the gains we obtained at the CE stage also carry over to the sMBR stage. Since the labeled dataset is much smaller, we use the GTC trainer with 16 GPUs for all models. From Table~\ref{tab:tab_sequence}, compared to a fully supervised CE trained AM, sMBR training yields a 10.7\% relative improvement in WER. In comparison to this sMBR model, we now compare two SSL student models on which sMBR training is performed, i.e., GTC and BMUF trained SSL student models. Sequence training only on labeled data still gives a good gain for SSL students (WER reductions of 18.6\% and 15.6\%, respectively over baseline labeled CE model), translating still into relative WER reductions of 8.2\% and 5.4\%, respectively over the fully supervised sMBR model. It is interesting to note that the effect of training with BMUF using 64 GPUs still does not fully recover after sMBR training with the GTC trainer using 16 GPUs, but we select this option for speed.

\section{Results on 1 Million Hours}
\label{sec:final_results}

In this section we present our final 1 Million hour model. We compare this model against the fully supervised sMBR model on (a) the acoustically difficult test sets with five device positions DP1 to DP5, and (b) TEST test set, along with sub-dividing it in two dimensions: nativity (TST-NATIVE, TST-NON-NAT) and SNR levels. We present these results in Tables~\ref{tab:tab_final_1},~\ref{tab:tab_final_2}.

\subsection{Training Convergence}
\label{sec:final_results_dist_train}
\begin{figure}[ht]
\begin{tikzpicture}

\begin{axis}[
  ylabel={Relative WER reduction ($\%$)},
  /pgfplots/ylabel shift=-5pt,
  xlabel={Sub-Epoch},
  y dir=reverse,
  xmin=1,
  xtick distance=2,
  ytick distance=5,
  height=5cm,
  width=9cm,
]
\addplot table {reg-lstm.dat};
\end{axis}
\end{tikzpicture}
\caption{\textit{On VAL test set, relative WER reduction per sub-epoch of the 1 Million hour SSL model against a baseline LSTM AM that is trained with CE criterion on the fully labeled 7,000 hour training data. Each sub-epoch corresponds to about 55,000 hours of data.}}
\label{fig:fig_1mhr}
\end{figure}
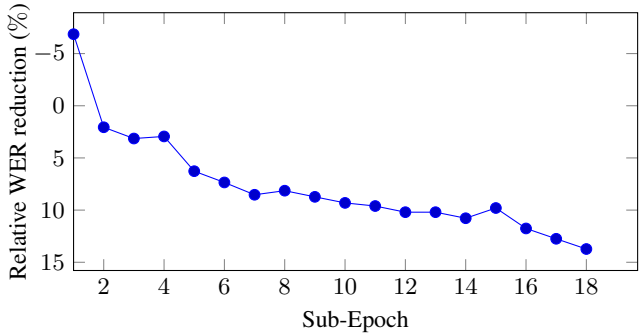

For the final 1 Million hour semi-supervised training we are using BMUF with 64 GPUs, using sMBR trained teacher, and employing scheduled learning. Figure~\ref{fig:fig_1mhr} plots convergence as WER reduction on the VAL set, relative to an LSTM AM that is trained with CE criterion on the fully labeled 7,000 hour training data with GTC trainer. The x-axis represents sub-epochs (each sub-epoch is about 55,000 hours of data), adding up all the way up to 1 Million hours. It can be seen that the student model  keeps improving up to the 14\textsuperscript{th} sub-epoch (i.e. up to 770,000 hours of data). Sub-epochs 16 to 18 are fine-tuning epochs, and the gains are larger. Note that the decrease in WER is not monotonic (sometimes there is even a slight increase), and we have not extensively tuned the learning hyper-parameters. From Figure~\ref{fig:fig_1mhr}, the WER reduction after training on the full 1 Million hours, at the CE stage, is 13.7\% -- significantly better than the corresponding 100,000 hour result (10.8\%) in Table~\ref{tab:tab_scheduled}.

\vspace{4mm}

\subsection{Final Results}

The final results including the final sMBR training can be seen in Table~\ref{tab:tab_final_1} and Table~\ref{tab:tab_final_2}. Except for the easiest device position (DP1), and the easiest noise condition (SNR$>$25dB), relative WER reductions are all greater than 10\%, and consistently the improvement is greater for harder conditions. Note also that the improvement is greater for non-native speakers. We take this as validation that large scale SSL can not only significantly improve accuracy overall (11.6\% error reduction), but also yield an out-sized improvement for the most challenging conditions.

\begin{table}[ht]
\centering
\caption{\textit{On the acoustically difficult test set (in DP1 to DP5), relative WER reduction ($\%$) of the final 1 Million hour model against a baseline LSTM AM that is sMBR trained on the fully labeled 7,000 hour training data.}}
\renewcommand{\arraystretch}{1.3}
\begin{tabular}{|c|c|c|c|c|c|}
  \hline
Test Conditions & WERR ($\%$) \\
 \hline
DP1 & $9.8$  \\
\hline
DP2 & $22.2$ \\
\hline
DP3 & $21.8$  \\
\hline
DP4 & $16.5$   \\
\hline
DP5 & $18.9$   \\
\hline
\end{tabular}
\label{tab:tab_final_1}
\end{table}

\begin{table}[ht]
\vspace{3mm}
\centering
\caption{\textit{On TEST test set, relative WER reduction ($\%$) of the final 1 Million hour model against a baseline LSTM AM that is sMBR trained on the fully labeled 7,000 hour training data.}}
\renewcommand{\arraystretch}{1.3}
\begin{tabular}{|c|c|c|c|c|c|}
  \hline
Test Conditions        & WERR ($\%$) \\
\hline
TEST & $11.6$   \\
\hline
\hline
TST-NATIVE & $11.6$   \\
\hline
TST-NON-NAT & $13.0$   \\
\hline
\hline
{TEST, SNR: $<$5 dB} & $13.3$   \\
\hline
{TEST, SNR: 5-10 dB} & $14.5$   \\
\hline
{TEST, SNR: 10-15 dB} & $10.7$   \\
\hline
{TEST, SNR: 15-20 dB} & $11.2$   \\
\hline
{TEST, SNR: 20-25 dB} & $12.9$   \\
\hline
{TEST, SNR: $>$25 dB} & $6.7$   \\
\hline
\end{tabular}
\label{tab:tab_final_2}
\end{table}

\section{Conclusions}
\label{sec:conclusions}
This paper reported on our lessons learned in building acoustic models on 1 Million hours of unlabeled speech data, in conjunction with 7,000 hours of labeled data. Using student-teacher learning, we simplified target generation without the need for decoding and confidence modeling. To optimize storage and to parallelize the target generation, we stored high valued logits from the teacher model. We introduced the notion of scheduled learning, interleaving learning on unlabeled and labeled data. This approach gave gains with CE and sMBR trained teacher models, but yielded bigger WER gains for CE trained teacher models. To scale distributed training to 64 GPUs we used BMUF, while performing sequence training only on the labeled data using GTC training with 16 GPUs. Our experiments showed that extremely large amounts of data are indeed useful; with little hyper-parameter tuning, we obtained relative WER improvements in the 10 to 20$\%$ range, with much higher gains in more difficult conditions, acoustically or in terms of speakers.

{\section*{\fontsize{9}{12}\selectfont Acknowledgements}\par}
We would like to thank Nitin Sivakrishnan for help with speeding up the data pipeline, Pranav Ladkat for implementing BMUF, Gautham Kollu for implementing an optimized forward propagation, Xing Fan for providing the setup for baseline models, and Harish Mallidi for help with the decoding infrastructure.

\bibliographystyle{IEEEbib}
\footnotesize
\bibliography{1mhr_ssl}

\end{document}